# A COMPUTATIONAL SYSTEM TO HANDLE THE ORTHOGRAPHIC LAYER OF TAJWĪD IN CONTEMPORARY QURʾĀNIC ORTHOGRAPHY

*Alicia González Martínez*

Contemporary Qurʾānic Orthography (CQO) relies on a precise system of phonetic notation that can be traced back to the early stages of Islam, when the Qurʾān was mainly oral in nature and the first written renderings of it served as memory aids for this oral tradition. The early systems of diacritical marks created on top of the Qurʾānic Consonantal Text (QCT) motivated the creation and further development of a fine-grained system of phonetic notation that represented tajwīd—the rules of recitation. We explored the systematicity of the rules of tajwīd, as they are encountered in the Cairo Qurʾān, using a fully and accurately encoded digital edition of the Qurʾānic text. For this purpose, we developed a python module that can remove or add the orthographic layer of tajwid from a Qurʾānic text in CQO[1]. The interesting characteristic of these two sets of rules is that they address the complete Qurʾānic text of the Cairo Qurʾān, so they can be used as precise witnesses to study its phonetic and prosodic processes. From a computational point of view, the text of the Cairo Quran can be used as a linchpin to align and compare Qurʾānic manuscripts, due to its richness and completeness. This will let us create a very powerful framework to work with the Arabic script, not just within an isolated text, but automatically exploring a specific textual phenomenon in other connected manuscripts. Having all the texts mapped among each other can serve as a powerful tool to study the nature of the notation systems of diacritics added to the consonantal skeleton.

[1] The code can be accessed at https://github.com/kabikaj/tajweed. A static html representation of the output data can be accessed at https://kabikaj.github.io/tajweed/.

## Introduction

The Cairo Qurʾān is currently the only complete Qurʾānic text fully and accurately encoded in digital form. This makes it an excellent choice to use it as a reference text for mapping other Quranic texts, in order to study and compare them by means of computational methods. Hence the importance of having a complete understanding and neat analysis of the Contemporary Qurʾānic Orthography (CQO) and precisely differentiate the layers in the script.

The phonetic component of the CQO and its relationship with tajwīd—the rules governing recitation—has been subject to multiple analyses in order to create computational systems that aim to enhance pedagogical practices. However, to our knowledge there has not been an effort to study the phonetic nature of tajwīd in relation to the writing system by means of a computational analysis. As we have already pointed out, the only Qurʾānic text for which we have currently a fully digitally encoded version is the Cairo Qurʾān. Thus, we can choose it as our target text for implementing a system of tajwīd rules governing the CQO.

## 1.1. A Qurʾānic Text

Initially, the Qurʾānic revelation was recorded orally and very early on accompanied with written artefacts that served as a memory aid to this oral tradition (Versteegh 2001, 55). Gradually, these written artefacts replaced the oral tradition as the main tool for preserving the revelation. This led to the need to rely on a phonetically accurate writing system that encoded the oral language, in a precise way but also open to record variability. For this purpose, Islamic scholars developed over time different systems of diacritical marks that served functions of diverse nature and used different shapes across time. Some of them, having supposedly the same purpose, were even used concurrently at times. Broadly speaking, one type of diacritical mark was used to disambiguate consonants that were written with the same archigrapheme, while the other type was used to indicate mainly vowels, both allowing the use of overlapping contradictory diacritics. These two types of systems were not used all over the base consonantal skeleton from the start but scattered through it. This complex and fluid orthographic and textual landscape, in addition to the divergences that early Qurʾānic manuscripts show among them, let us identify different readings of the text, i.e., different qirāʾāt[2] قراءات, which constituted an essential tool used to standardise the Qurʾānic text (Nasser 2020, chap. 1, 2). It may be worth mentioning in this connection that the Prophet is said to have recited the same passage in various ways at various times (Graham 1987, 99)

Among the available canonical readings of the Qurʾān, the Ḥafṣ' ʿan ʿĀṣim reading has turned to be the most widespread Qurʾānic text used in contemporary times as it was used as the model, under the supervision of Al-Azhar (Puin 2009, 606; Nasser 2020, chap. 1, 8), for the Cairo printed edition of the Qurʾān[3], in 1923-1924[4]. Other readings are still in used in some parts of the world—for example Warš ʿan Nāfiʿ is still widely used in Morocco, Algeria and Tunisia[5] (van Putten, 2022, 5-6). Subsequently, the Cairo Edition was used as the most common model for the digital encoding of the Qurʾānic text. The Cairo Qurʾān reflects the CQO, which diverges from Modern Standard Arabic (MSA) and the ʿUthmānic text (UT). For instance, CQO makes use of miniature letters inserted amphibiously—i.e., with a horizontal offset vis-à-vis the carrier grapheme—above the Qurʾānic Consonantal Text (QCT) (Milo 2009, 492), whereas MSA and UT don't show this trait.

---

[2]We use the DIN 31635 standard to transcribe Arabic script throughout the article.

[3]As Graham mentions (Graham 1987, 55), the text was not based on available manuscripts but upon the oral and written traditions recorded in the qirāʾāt literature, following the Ḥafṣ ʿan ʿĀṣim reading.

[4]Over the literature, there seems to be some confusion on the exact year of publication (Graham 1987, 211). Although the year indicated in the print of this first edition was 1919, the printing itself took several years to be finished (Puin 2009, 606).

[5]Abuhamdia (2016) mentions that Nāfiʿ is in general followed in the Maghreb and in West African Muslim countries, Bin Al-ʿAlāʾ in Sudan and parts of Yemen, and ʿĀṣim in Afghanistan, India, Pakistan, Turkey and the rest of the Arab countries.

## 1.2. Previous Work

There have been several works on tajwīd that included a computational component. As we already pointed out, all of them aim at enhancing pedagogical needs. The most recent studies focus on speech recognition (Ahmad et al., 2018; Alagrami and Eljazzar, 2020), hence not relevant for the present work. Among the projects that involve text, we can mention two them:

Alfaries et al. (2013) developed a tool called *QurTaj* for automatically annotating tajwīd by describing contextual rules for each type of tajwīd phenomenon. The rules are defined on top of a tokeniser that splits the Qurʾānic text into words, letters and diacritics. Unfortunately, the system is only evaluated against the surah Al-Šuʿarāʾ الشعراء, which means that only 1.7% of the whole Qurʾānic text is checked.

Aqel and Zaitoun (2015) describe an expert system for helping non-Arabic Muslim speakers to identify tajwīd rules. The system asks the user several questions about the orthographic context of the sequence and based on the results it identifies the rule. An evaluation to the system is mentioned by the authors, but no details about it are given.

## Digital Qurʾāns

Having a complete and accurate encoding of the Qurʾānic text allows us to explore in a systematic way the rules governing tajwīd in CQO. To our knowledge, only two projects have created digital Qurʾānic texts that made use of CQO: The Tanzil Qurʾān and the DecoType Qurʾān.

Both Qurʾāns are actually based on the King Fahd Complex for the Printing of the Holy Qurʾān edition (1985), also known as the Medina Qurʾān. But this in turn is based on the second edition of the King Fuʾād Qurʾān, i.e., the Cairo Edition, published in 1952. According to Puin (2009, 602), the difference between the two is that the Medina Qurʾān is based on a handwritten copy made by the calligrapher ʿUṯmān Ṭaḥa عثمان طه, whereas the Cairo Qurʾān was printed using movable type, hence the deviation of some calligraphic rules in this version compared to the Medina one[6].

[6]From the point of view of the context, according to Philipp Bruckmayar, the Medina Edition is actually a complete plagiarism of the Cairo Edition, except for two letters (https://www.ideo-cairo.org/2021/12/the-cairo-edition-of-the-qur%CA%Bean-1924-texts-history-challenges, visited 30 March 2023)

### 1.3.1. The Tanzil Qurʾān

The Tanzil project (https://tanzil.net) was created in 2007 with the aim to produce a digital Qurʾānic text that followed a reliable Unicode encoding and that was freely available for non-commercial purposes. The text is provided in different orthographic versions, including one in a complete ʿuṯmānī script. Although they provide a fairly accurate encoding of the ʿuṯmānī text, there are minor inconsistencies[7] and, more importantly do not use open tanwīn diacritics, i.e., U+08f0, U+08f1 and U+08f0, which are a type of tajwīd marking. For instance, Q2:90:26 is encoded as عَذَابٌ instead of عَذَابࣰ. Already this sole omission makes the Tanzil Qurʾān not suitable for the present study.

### 1.3.2. The DecoType Qurʾān

The DecoType Qurʾān was used to make the Muṣḥaf Muscat [https://mushafmuscat.om], a digital Qurʾān commissioned by the ministry of Endowments and Religious Affairs of the Sultanate of Oman and contracted out to the Dutch company DecoType. Its website was launched on 12th June 2017 (Milo 2022, 25). DecoType ignores conventional font technologies, which are based on legacy thinking and are constrained by the restricted technological capabilities of the printing press. Instead, they created a novel technology for digital typography—called the Advanced Composition Engine (ACE)—that makes use of the full potential of computers and that is capable of accurately displaying any Arabic script with absolute precision in an elegant and scalable way. By doing so, they created "arguably the most advanced typographical technology in the world" (Nemeth 2017, 410). As a result, the Muṣḥaf Muscat displays Arabic script as it was attested in actual manuscripts (van Lit 2020, 190-191), contrary to the ineffective movable types of the Cairo Edition, and in line with the *calligraphic* Medina Qurʾān.

[7] These subtle encoding inconsistencies are normally handled properly by the rendering technologies, but the fact that they fail to provide a completely accurate encoding underneath poses a problematic challenge to studies aiming to explore in fine detail elements of the Arabic writing system, such as the present one.

An interesting and much desired side effect of using DecoType's rendering technology is that they don't need to resort to any *hacks* in the way they encode Arabic text, since they rely on a linguistically consistent separation of script encoding from script rendering. For example, Tanzil Qurʾān encodes Q56:36:1 as فَجَعَلْنَـٰهُنَّ◌, so using Arabic taṭwīl (U+0640), clearly to *help* font technologies display the word correctly. On the contrary, DecoType encodes it as فَجَعَلْنَٰهُنَّ, so without taṭwīl, leaving display decisions to ACE.

Though encoding practices as the one described before can easily be overcome by normalising the text, the lack of open tanwīn characters prevents us from studying tajwīd automatically. Fortunately, DecoType includes them. Therefore, we chose this Qurʾānic text for our study[8].

## Tajwīd تجويد

Tajwīd is the system of rules regulating the correct and clear oral rendering of the Qurʾān, in order to preserve the revelation, both in terms of its sounds, as well as its content (Nelson 2001, 14; 2006, Vol. 4, 425). This consists of rules that govern how phonemes assimilate to their context, both within words and across word or morpheme boundaries. Additionally, it includes rules for vowel lengthening and guidelines for pauses during speech, which mostly serve semantic purposes. In this sense, tajwīd rules play a central role in prosody and their study can be highly valuable for understanding how the latter works (Bohas et al. 1990, 96).

Below, we list partially the tajwīd rules in a non-comprehensive way, loosely following the description presented by Czerepinsky (2012), but mostly focusing on those rules that affect script and, sometimes, oversimplifying or excluding elements not directly relevant to the present study. We are not including Qurʾānic stops (ۛ ج لا م قلى صلى) because we cannot make any contextual rules for them.

---

[8]I would like to thank Thomas Milo for kindly giving me permission to use the DecoType Qurʾān for making the present study and for all his valuable insights on Unicode and CQO.

# 1. al-nūn al-sākinah النون الساكنة

Nūn not bearing any vowel-diacritic (ḥarakāt)—either with sukūn or without anything. This also includes the tanwīn mark of the indefinite article, which contains a *hidden* nūn. The nūn changes or not its pronunciation according to the following consonant:

## 1.1. al-iẓhār الإظهار

This consists of pronouncing every letter according to its articulation point. This happens when the nūn is followed by ء /ʾ/, ه /h/, ع /ʿ/, خ /ḫ/, ح /ḥ/ or غ /ġ/, which are conventionally called *guttural* letters (الحروف الحلقية)—which correspond to laryngeals, pharyngeals, and uvular fricatives (Watson 2002, 37). As there are no sound changes in this context, the orthography is the default one. An example at word boundary is مِنْ خَشْيَةِ Q2:74:30-31, and in internal position صَنْعَةَ Q21:80:2.

## 1.2. idġām الإدغام

This refers to consonantal regressive assimilation (Abuhamdia 2016), and can be of various types:

### 1.2.1. idġām bi-ġunnah إدغام بغُنّة

It the nūn is assimilated by the following phoneme causing nasalisation (ġunnah الغُنّة).

### 1.2.1.1. idġām bi-ġunnah kāmil إدغام بغُنّة كامل

This is caused when the nūn is followed by م /m/, ن /n/. As a result of the rule, the second phoneme of the cluster is geminated. Orthographically, this is marked by not writing a sukūn on the first consonant and adding a šaddah on the second consonant. In the case of tanwīn, for fatḥah and kasrah, the two ḥarakāt are not placed exactly one above the other, but they are just partially overlapping, whereas in the case of ḍammatan, instead of the conventional ligature, the ḍamma diacritics are written separately side-by-side. Examples of this rule are وَلۡتَكُن مِّنكُمۡ Q3:104:1-2 and لُوطٍ نَّجَّيۡنَٰهُ Q54:34:7-8. There is no case of silent nūn followed by mīm in internal word position. As for the nūn followed by nūn, it's marked with šaddah, as to be expected.

### 1.2.1.2. idġām bi-ġunnah nāqiṣ إدغام بغُنّة ناقص

This is caused when the nūn is followed by ي /y/ or و /w/. The first consonant of the cluster, i.e. the /n/, is assimilated into the glide, which retains a nasalisation. As the /n/ is not completely lost, the assimilation is not considered "complete", so no šaddah is added on the glide consonant. The only mark of this assimilation is the absence of the sukūn on the nūn or the displaced tanwīn. For example, فَلۡيُؤۡمِن وَمَن Q18:29:7-8 and حَسَنَةً يُضَٰعِفۡهَا Q4:40:9-10.

There is no idġām within a word, only across word boundaries. We checked all the cases and found the 4 lemmas typically mentioned in the literature: بُنۡيَٰنٌ، صِنۡوَانٌ، قِنۡوَانٌ، ٱلدُّنۡيَا. They make 10 different word forms and a total of 125 occurrences.

### 1.2.2. idġām bi-ġayr ġunnah إدغام بغير غُنّة

This occurs when the *silent* nūn is followed by /l/ ل or /r/ ر. In this case, the nasalisation is completely lost in the assimilation and the liquid consonant geminated as a result. As in the case of the idġām bi-ġunnah kāmil, the first consonant of the cluster is stripped of the sukūn—or we find a displaced tanwīn—and the second letter is reinforced with a šaddah. For example, يَكُن لَّهُۥ Q112:4:2-3 and رَّحِيمٌ رَّبَّنَآ Q14:36:15-37:1.
Czerepinsky points to an exception, مَنْۜ رَاقٍ Q75:27:2-3, where a miniature sīn indicates that a pause (sakt) should be made between the two. The function of this pause is probably to emphasize the personal pronoun, so it is done for the sake of semantics.

## 1.3. al-qalb القَلْب

This happens when the vowelless nūn is followed by /b/ ب. In this case the /n/ changes its place of articulation to that of the /b/, turning into a /m/. Orthographically, the nūn carries a miniature mīm instead of a sukūn, and the tanwīn is done with a single ḥarakah plus a miniature mīm (below in the case of a kasrah). Examples are مِنۢ بَعْدِهِمْ Q7:169:2-3, فَأَنۢبَتْنَا Q.80,27,1 and غُرْفَةًۢ بِيَدِهِۦ Q2:249:23-24.

## 1.4. al-ʾiḫfāʾ الإخفاء

A consonant that is not in any of the groups described above—i.e., not a guttural, liquid, glide nor nasal—when preceded by vowelless nūn, turns it into an approximant, i.e., the /n/ is assimilated into the next consonant leaving a nasalization. The list of consonants are /t/ ت, /ṯ/ ث, /ǧ/ ج, /d/ د, /ḏ/ ذ, /z/ ز, /s/ س, /š/ ش, /ṣ/ ص, /ḍ/ ض, /ṭ/ ط, /ẓ/ ظ, /f/ ف, /q/ ق, /k/ ك. As the /n/ is not completely lost, orthographically this rule is marked with the absence of sukūn above the nūn or with the displaced tanwīn. This rule works both in internal word position and at word boundary. For example, مِن شَرِّ Q113:2:1-2, هَٰٓأَنتُمْ Q4:109:1 and مُّسَمًّى ثُمَّ Q22:33:6-7.

## 2. al-mīm al-sākinah الميم الساكنة

As happens with the nūn, the mīm may assimilate or not depending on the neighbouring consonants.

### 2.1. al-ʾiḫfāʾ al-šafawiy الإخ الشفوي

Non-vowelled /m/ followed by /b/. The mīm turns into an approximant and this is marked by the absence of sukūn. There are no cases inside a word. An example of the rule is هُم بِرَبِّهِمْ Q23:59:2-3.

### 2.2. al-idġām al-miṯlayn الإدغام المثلين

Non-vowelled /m/ follows by /m/. There is complete assimilation, so the first mīm does not bear sukūn and the second mīm has a šaddah. This context within a word is resolved with a single mīm with šaddah. An example of this rule is لَهُم مِّن Q7:41:1-2.

### 2.3. al-iẓhār al-šafawiy الإظهار الشفَوي

There is no assimilation and therefore no orthographic marking. This happens with the rest of the consonants. For example, أَلَمْ تَعْلَمْ Q2:107:1-2.

## 3. al-lām al-sākinah اللام الساكنة

This happens when the vowelless lam of the definite article precedes a coronal consonant[9]. In this context complete assimilation is supposed to occurs, and the /l/ is totally lost, and the coronal consonant is geminated. Orthographically, as in previous cases, the lām does not bear sukūn and the following consonants has a šaddah. This *silent* lām also occur when a vowelless lām is followed by ل /l/ or ر /r/, either between word boundary or not. Examples ٱلتَّوْرَىٰةَ Q3:3:10 and بَل رَّفَعَهُ Q4:158:1-2.

There is an exception to this assimilation rule in بَلْ رَانَ Q83:14:2. As before, this context seems to be semantically motivated.

## 4. al-madd المدّ

From a very general perspective, the rules of al-madd refer to different types of lengthening of long vowels—in terms of dura-

[9] The letter jīm, typically pronounced as a voiced palato-alveolar affricate, though part of this class, probably evolved from an original velar articulation *g, hence not assimilating in presence of the lām of the definite article (Watson 2002, 210).

tion—according to their context, so it is a sign indicating an overlong vowel[10] (van Putten 2021). We include here, not just the maddah sign per se, i.e. the diacritic ◌ٓ, but also the miniature waw ۥ and the miniature ya ۦ.

## 4.1. al-madd al-wāǧib al-muttaṣil المدّ الواجب المتّصل

It occurs when a long vowel is followed by a hamza. It in case, a madd diacritic is added on top of the so called *madd letter*. For example, يَشَآءُ Q2:90:18.

## 4.2. al-madd al-ǧāʾiz al-munfaṣil المدّ الجائز المنفصل

This rule describes the same context of the previous letter, but when ocurring at word boundary. For example, بِهِۦٓ إِلَّا Q83:12:3-4.

## 4.3. al-madd al-lāzim المدّ اللازم

Simplifying, this context is basically met when a long vowel is followed by a geminated consonant. For example, ٱلْحَآقَّةُ Q69:1:5. There is also al-madd al-lāzim on top of some of the letters that appear at the beginning of some Qurʾānic surahs. For example, طسٓمٓ Q26:1:5.

[10] For a precise description of the use of maddah in CQO see Milo (2009, 507)

### 4.4. al-madd al-ṣilah المدّ الصلة

The enclitic pronoun of third person masculine singular -ه, or the ha of the feminine demonstrative singular هذه, when preceded by a vowel, is lengthened with a miniature waw or ya, for ḍamma and kasra respectively, when it is not followed by hamza from the next word. For example, أَهْلَهُۥ Q2:126:10.

There are two exceptions to this rule: (1) يَرْضَهُ Q39:7:13 is not lengthened even though it meets the described context, and (2) فِيهِۦ Q.25:69:7 is lengthened though not adjusting to the context.

## 5. al-idġām for Consonants other than Nūn and Mīm

When a vowelless consonant is followed by the same consonant—as it happened in the case of al-idġām al-miṯlayn with the mīm letter—their assimilation is marked by the absense of sukun in the first one a a shadda in the second. For example, يُدْرِككُّمُ Q4:78:3. On the other hand, a consonant cluster of two letters that are not identical but similar in their place or manner of articulation can also cause assimilation and it's marked accordingly. For example, نَخْلُقكُّم Q77:20:2.

## 6. al-ṣifr al-mustadīr الصفر المُستَدير

A diacritic with the shape of a small high rounded zero—encoded in Unicode as U + 06df—that is placed on top of some letters alif, waw and ya, indicate that these letters should not be pronounced neither in connection nor pause. For example, أَجْرَمُوا۟ Q6:124:21.

### 7. al-ṣifr al-mustaṭīl al-qā'im الصفر المستطيل القائم

A diacritic with the shape of a small high upright rectangular zero—encoded in Unicode as U+06e0—on top of an alif indicates that the alif is additional in consecutive reading but should be pronounced in pause. For example, لَّٰكِنَّا۠ Q18:38:1

### 8. al-sakt السكت

The al-sakt is a gentle pause made without breathing while reciting. It is marked by a miniature sīn. We have few examples of this rule, and we need to consider them lexically, for example, the case seen below in مَنۜ رَاقٍ Q75:27:2-3, that was blocking a rule of idġām.

## *Tajwīdiser*: a System of Orthographic Rewrite Rules

The system we created for adding and removing the tajwīd orthographic layer can be found in github at github.com/kabikaj/tajweed. A static html visualisation of the data can be found at kabikaj.github.io/tajweed. The script containing all the rules is called tajweed.py. The rules are written in python using regular expressions. It is worth mentioning that there is not necessarily a one-to-one relationship with a regular expression rule and an orthographic rule. The rules are applied either within words, across word boundaries or on both.

We run a two-phase process on the Qur’ānic text. First, the tajwīdiser applies a set of cascade rewrite rules to remove the tajwīd and restore sukūn diacritics when required. Then, another set of reverse cascade rules are applied to the detajwīdised text to add back the tajwīd layer. At the end two types of verifications are conducted: (1) that the original text and the restored tajwīdised text match, and (2) that there are no missing marks of tajwīd in the detajwīdised text. In this way, we can assure the effectiveness and accuracy of our rules.

From an overall perspective, we have three main types of rules: assimilation rules, elongation rules and pausal marks.

## 1. Assimilation rules

this group includes al-nūn al-sākinah (in the code, rules with ids 'N2.1.1.A', 'N2.1.1.B', 'N2.1.1.C', 'N2.1.1.D', 'N2.1.2.A', 'N2.1.2.B', 'N2.1.2.C', 'N2.1.2.D', 'N2.2.A', 'N2.2.B', 'N2.2.C', 'N2.2.D', 'N3.A', 'N3.B', 'N3.C', 'N3.D', 'N4.A', 'N4.B', 'N4.C', 'N4.D'), al-mīm al-sākinah (ids 'M1', 'M2'), al-lām al-sākinah of the definite article (id 'SHAMS'), as well as other assimilation processes (idġām) affecting consonant clusters (ids with rules 'MITHL-bb', 'MITHL-dd', 'MITHL-kk', 'MITHL-ll', 'MITHL-yy', 'MITHL-hh', 'MITHL-ww', 'MITHL-tt', 'MITHL-rr', 'MITHL-ðð', 'MITHL-ff', 'MITHL-33', 'MTJNS-dt', 'MTJNS-td', 'MTJNS-tT', 'MTJNS-Tt', 'MTJNS-ṯð', 'MTJNS-lr', 'MTJNS-ðṮ', 'MTJNS-qk', 'MTJNS-bm', 't-assim').

It is interesting to point out that, in order to describe the lām sākinah rule affecting the definite article, we needed to count with the part-of-speech of all the words of the Qurʾānic text, as the rule only affects nouns—or more properly speaking, ism in the Arabic grammatical tradition, i.e., words that are not verbs فعل fiʿl or particles حرف ḥarf. Indeed, if we apply the rule for the assimilation of the article to all words, we over generate and apply the rule to cases where the assimilation doesn't occur. To solve this problem, we used the morphosyntactic analyses provided by the corpus.quran.com project (Dukes and Habash 2010. We processed their data and mapped its linguistic information with our Qurʾānic text. After that, we indicated in our assimilation rule for the definite article that it should only be applied to words tagged as noun. The resulting reverse rule for removing and adding the orthographic layer of this rule is flawless, covering all cases.

Apart from the assimilation in the definite article, if we pay a closer look at the assimilations occurring on consonant clusters, regardless being at word boundary or internally, all of them cause the first consonant to lose its sukun. Then, if the assimilation is complete, i.e., the sound of the first consonant is totally lost, then the second consonant bears a shadda. On the other hand, if there are traces of the manner or place of articulation of the first consonant, no shadda is added into the second.

Obviously, the presence of otiose alifs does not prevent the assimilations to take place, for example عَصَواْ وَّكَانُواْ Q2:61:57-8.

Apart from consonant clusters containing the same repeated phoneme, these are the contexts that cause complete or partial assimilation:

/n/ followed by any consonant that is not a guttural

/m/ followed by a /b/ or /n/. The first, /mb/, only occurs at word boundary. The sequence /b/ followed by /m/ only occur once, at word boundary, and it also assimilates: ٱرۡكَب مَّعَنَا Q11:42:14-15.

/d/ followed by /t/: both at word boundary and in internal position. There are 45 cases. The reverse, /t/ followed by /d/, also occurs and assimilate, but there are only 2 cases ad both at word boundary: أَثۡقَلَت دَّعَوَا Q7:189:21-22 and أُجِيبَت دَّعۡوَتُكُمَا Q10:89:3-4.

/t/ followed by /ṭ/: at word boundary, 14 cases. There are no cases in internal position. The reverse, /ṭ/ followed by /t/ only occurs in internal position 4 times (Q5:28:2, Q12:80:21, Q27:22:5 and Q39:56:7), and it also assimilate.

/ṯ/ followed by /ḏ/: at word boundary, only one case: يَلۡهَثۚ ذَّٰلِكَ Q7:176:20-21. There are no cases in internal position.

/l/ followed by /r/: operative at word boundary, there are 12 cases in total, and no cases in internal position.

/ḏ/ followed by /ẓ/: only two cases and both at word boundary: إِذ ظَّلَمُوٓا۟ Q4:64:11-12

/q/ followed by /k/: only one case, which is in internal position: نَخۡلُقكُّم Q77:20:2

## 2. Elongation rules

This group includes the madd rules containing the madd sign on top of alif, waw or ya (rule ids 'MADD-hmz', 'MADD-hmz-A-sil', 'MADD-hmz-sp-1, 'MADD-hmz-sp-2, 'MADD-hmz-sp-3, 'MADD-hmz-sp-4, 'MADD-lzm', 'MADD-shdd-skn', 'MADD-sp-1', 'MADD-sp-2', 'MADD-sp-3', 'MADD-sp-4', 'MADD-sp-5', 'MADD-sp-6', 'MADD-sp-7', 'MADD-sp-8', 'MADD-sp-9', 'MADD-sp-A', 'MADD-sp-B', 'MADD-sp-C', and 'MADD-sp-D'), the miniature waw (ids 'HU', 'min-u-1', 'min-u-2', 'min-u-3', 'min-u-4', 'min-u-5', 'min-u-6', 'min-u-7', 'min-u-8') and the miniature ya (ids 'HI', 'min-y-1', 'min-y-2', 'min-y-3', 'min-y-4', 'min-y-5', 'min-y-6', 'min-y-7', 'min-y-8', 'min-y-9', 'min-y-A').

The main rule for the elongation before a hamza is 'MADD-hmz', so it corresponds to al-madd al-wāǧib al-muttaṣil and al-madd al-ǧāʾiz al-munfaṣil. As for the madd before a geminated, i.e., the al-madd al-lāzim, the rules are 'MADD-lzm' and 'MADD-shdd-skn'.

The rules 'HU' and 'HI' define the elongation of the ḍamma and kasra of the enclitic pronoun, i.e., al-madd al-ṣilah. As stated in the literature, there is only one exeption to this rule, which we had to indicate: يَرۡضَهُ Q39:7:13.

The rest of the rules indicated in this group are lexical, i.e., they correspond to specific word forms or lemmas, and thus would require a more detailed description that is out of the scope of this analysis.

## 3. Pausal rules

The pausal marks included in this study include three types: al-ṣifr al-mustadīr (rule ids 'Sil-1', 'Sil-2', 'Sil-3', 'Sil-4', 'Sil-5', 'Sil-6', 'Sil-7', 'Sil-8', 'Sil-9', 'Sil-A', 'Sil-B', 'Sil-C', 'Sil-D', 'Sil-E', 'Sil-F', 'Sil-G'), al-ṣifr al-mustaṭīl al-qāʾim (ids 'P-sil-1', 'P-sil-2'), and al-sakt (ids 'sakt-1', 'sakt-2', 'sakt-3').

As already mentioned, it is important to note that some rules are applied only to specific word forms. This is especially true for the pausal rules. In some cases, this might mean that we didn’t succeed completely in defining the context of a rule, so we needed to resort to combine the rule with a list of exceptions or restrictions. But in most of the cases, restricting a rule through a list of word forms indicates that the rule is related with semantics and not with the phonetic context. For example, the same word is repeated twice with different pausal marks at the end, first as قَوَارِيرَا۠ Q76:15:8 and then as قَوَارِيرَا۟ Q76:16:1, meaning that in the first one the alif should be read when making a pause, whereas in the second word the alif should not be read when stopping. Obviously, we had to treat this case lexically.

As regards the sakt sign, we decided to include it because it affects few cases, although we could have left it aside as in the case of the Qurʾānic Stops, since we cannot define rules related to the context where it appears.

## Discussion

A detailed study of the tajwīd orthographic layer based on the text of the Cairo Edition might not seem very interesting at first. The Fuʾād Muṣḥaf was not based on the study of the early Qurʾānic manuscripts still available, as it was thought that they reflected an imprecise script caused by the lack of competence of the scribes that crafted them (Puin 2011, 147). It might therefore seem that we should rather direct our efforts to produce critical editions of the earliest manuscripts. This is a valid point. But from the digital point of view, we can greatly benefit from having a complete and accurately encoded reference Qurʾānic text enriched with different layers of linguistic information that can be used as a linchpin for aligning the different Qurʾānic manuscripts and all phenomena described in the literature. Instead of producing disconnected and static digital editions, we should make use of the full potential of computational technology by building dynamic and robust systems based on a combination of accurately encoded texts and linguist rules. By doing so, we can generate alternative texts straight away [11].

[11] This is inspired by the way the Advanced Composition Engine (ACE) deals with digital typography.

In the case of our study, we have implemented a set of rules that let us generate a Quranic text without tajwīd straight away.

In this scenario, adding a new rule to this system—for instance, a different set of assimilation rules for consonant clusters based on other readings than the one of Ḥafṣ' ʿan ʿĀṣim—is almost effortless. Furthermore, we can perform rigorous evaluations on this type of system, as in the case of this study, where we have proved that our two-phase conversion produces the same exact text as the original one.

The development of this type of system must satisfy the following desiderata:
- appropriate methodology directed to create dynamic systems of text encoding, instead of frozen and static representations of text,
- accurate digital encoding making a clear distinction between a linguistically based script encoding versus a rendering mechanism for visualisation.
- simple collation mechanisms,
- linguistically based rules that can be applied to the base texts and to which we can potentially apply well-informed modifications in order to create alternative texts.

## Conclusions

We have described and implemented a system for digitally handling the rules of tajwīd in CQO that addresses the full text of the Cairo Qurʾān. The rules of this system, which have been implemented using regular expressions, describe the contexts where each tajwīd phenomena must be applied, indicating possible exceptions and restrictions. All this information can be directly used to study more in depth the system of tajwīd rules governing the Ḥafṣ' ʿan ʿĀṣim reading according to the Cairo Qurʾān. More interestingly, this resulting Qurʾānic text enriched with an optional layer of tajwīd data could be used as a starting point to align and compare the different canonical readings of the Qurʾān, creating a mapping among the different manuscripts.

## Acknowledgments

This research was part of the InterSaME project, which was funded by the DFG under a DFG-AHRC Cooperation.

## References

Versteegh, Kees. 2001. *The Arabic Language*. Edimburgh: Edimburgh University Press.

Nasser, Shady. 2020. *The Second Canonization of the Qurʾān (324/936)*. Brill.

Graham, William A. 1987. *Beyond the Written Word: Oral Aspects of Scripture in the History of Religion*. Cambridge University Press.

Puin, Gerd. 2009. 'Quellen, Orthographie und Transkription moderner Drucke des Qurʾān'. In *Vom Koran zum Islam,* edited by Markus Groß and Karl-Heinz Ohlig. Verlag Hans Schiler

van Putten, Marijn. 2022. *Quranic Arabic. From its Hijazi Origins to its Classical.* Leiden: Studies in Semitic languages and linguistics. Leiden: Brill.

Milo, Thomas. 2009. 'Arabic Amphibious Characters. Phonetics, phonology, orthography, calligraphy and typography'. In *Vom Koran zum Islam*, edited by Markus Groß and Karl-Heinz Ohlig. Verlag Hans Schiler.

Ahmad, Fadzil, Yahya, Saiful Zaimy, Saad, Zuraidi, Ahmad, Abdul Rahim. 2018. 'Tajweed Classification Using Artificial Neural Network'. International Conference on Smart Communications and Networking (SmartNets).

Alagrami, Ali M., Eljazzar, Maged M. 2020. 'Smartajweed automatic recognition of Arabic Quranic recitation rules'. International Conference on Computer Science, Engineering And Applications

Abuhamdia, Zakaria Ahmad. 2016. 'Phonetically motivated and phonetically unmotivated assimilation in tajweed'. Journal of Philosophy, Culture and Religion. Volume 17.

Aqel, Musbah J., Zaitoun, Nida M. 2015. 'Tajweed: An Expert System for Holy Qur'an Recitation Proficiency'. Procedia Computer Science. Volume 65.

Milo, Thomas. 2022. *About Mushaf Muscat.* Georg Olms Verlag.

Nemeth, Titus. 2017. *Arabic Type-Making in the Machine Age. The Influence of Technology on the Form of Arabic Type, 1908–1993.* Leiden: Brill.

Van Lit, L.W.C. 2020. *Among Digitized Manuscripts. Philology, Codicology, Paleography in a Digital World.* Brill

Nelson, Kritina. 2001. *The art of reciting the Qurʾān*. The American University in Cairo Press.

Nelson, Kritina. 2006. 'Tajwīd'. In *Encyclopedia of Arabic Language and Linguistics* edited by Kees Versteegh. Volumen 4. Brill.

Bohas, Georges, Guillaume, Jean-Patrick, Kouloughlu, Djamel. 1990. *The Arabic Linguistic Tradition*. Washington, D.C: Georgetown University Press.

Czerepinsky, Kareema. 2012. *Tajweed rules of the Qur'an*. Kalamullah

Watson, Janet. 2002. The phonology and morphology of Arabic. Oxford University Press.

Van Putten, Marijn. 2021. 'Madd as orthoepy rather than orthography'. Journal of Islamic Manuscripts 12 (2021) 202–213.

Dukes, Kais, Habash, Nizar. 2010. Morphological annotation of Quranic Arabic. Proceedings of the Seventh International Conference on Language Resources and Evaluation (LREC'10).

Puin, Gerd. 2011. 'Vowel letters and ortho-epic writing in the Qurʾān'. In *New perspectives on the Qurʾān* by Gabriel Said Reynolds. Routledge.